\documentclass[runningheads]{llncs}

\usepackage{eccv}

\usepackage{eccvabbrv}

\usepackage{graphicx}

\usepackage{orcidlink}

\usepackage{enumitem}
\usepackage{amsmath}
\usepackage{amssymb}
\usepackage{bm}

\usepackage{booktabs} 
\usepackage{multirow}
\usepackage[table]{xcolor}
\definecolor{medgreen}{RGB}{235, 250, 235}
\definecolor{headergray}{gray}{0.93}
\definecolor{color4}{RGB}{220, 215, 238}
\definecolor{color3}{RGB}{215, 222, 242}
\definecolor{color2}{RGB}{210, 228, 250}
\definecolor{color1}{RGB}{208, 232, 228}

\usepackage{graphicx}

\begin{document}

\title{CoRE: Concept-Reasoning Expansion for Continual Brain Lesion Segmentation} 

\titlerunning{Abbreviated paper title}


\author{Qianqian Chen\inst{1} \and
Anglin Liu\inst{2} \and
Jingyang Zhang\inst{1,*} \and
Yudong Zhang\inst{1,*}}

\authorrunning{Q.~Chen et al.}

\institute{Southeast University, Nanjing, China \and
The Hong Kong University of Science and Technology (Guangzhou), Guangzhou, China\\
* Co-corresponding authors\\
\email{\{j.y.zhang, yudongzhang\}@seu.edu.cn}} 

\maketitle

\begin{abstract}
  Accurate brain lesion segmentation in MRI is vital for effective clinical diagnosis and treatment planning. Due to high annotation costs and strict data privacy regulations, universal models require employing Continual Learning (CL) to adapt to evolving clinical tasks without losing previously acquired knowledge. However, existing CL paradigms often suffer from capacity limits or redundant parameter growth, and even advanced dynamic methods rely mostly on image-perception strategies that struggle to handle the substantial pathological and multimodal heterogeneity inherent in brain imaging.
  To address this issue, we propose Concept-Reasoning Expansion (CoRE) framework, which establishes a joint decision-making mechanism by integrating visual features with structured concepts. 
  Through the alignment of image tokens with a hierarchical concept library, CoRE simulates clinical reasoning to guide both interpretable expert routing and demand-based model growth. This collaborative process ensures model evolution is grounded in clinical priors, preventing redundant parameter expansion while maximizing knowledge reuse. Extensive evaluations across 12 sequential brain lesion MRI tasks demonstrate that CoRE achieves state-of-the-art performance and provides a high knowledge starting point for efficient future adaptation. Its superior few-shot transferability and clinical interpretability further validate its effectiveness in managing non-stationary clinical data streams. Our code will be released soon.
  
\end{abstract}

\section{Introduction}
\label{sec:intro}
Accurate segmentation of brain lesions is crucial for diagnosing diseases and planning treatments\cite{kamnitsas2017efficient}. While deep learning has significantly advanced this field\cite{czolbe2023neuralizer,wood2022deep,basaran2023lesionmix}, the task remains challenging due to pathological heterogeneity and modality diversity\cite{wu2006characterizing,shah2013discriminating}. 
Therefore, prevailing methods often deploy specialized models tailored to specific lesion types or modalities~\cite{gabr2020brain,zhang2024foundation}, which scale poorly and are inefficient in clinical practice. To address this, developing universal models across varied tasks has emerged as a promising solution\cite{liu2023clip,liu2024universal}. However, training such models typically relies on joint learning, which requires the long-term retention of historical data that is severely restricted by strict privacy regulations and limited computational resources\cite{zhang2024challenges}. Hence, to build a universal brain lesion segmentation model under these constraints, the framework is highly desired to incorporate Continual Learning (CL) capabilities~\cite{10444954}. CL empowers the model to sequentially adapt to new tasks while preserving previously acquired knowledge without the need to access prior data, making it a critical paradigm for sustainable clinical deployment.

\begin{figure}[t!] 
    \centering     
    \includegraphics[width=\linewidth]{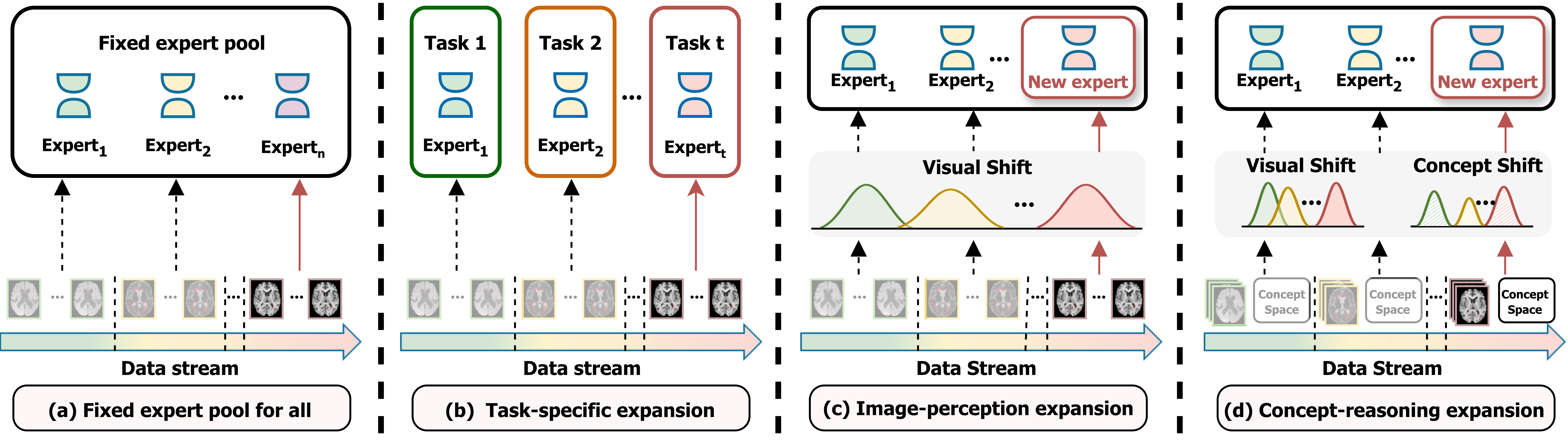} 
    \caption{Illustration of CL paradigms. \textbf{(a-b) Traditional paradigm (left)}: Existing methods typically \textbf{(a)} employ a fixed expert pool shared by all tasks or \textbf{(b)} add task-specific experts, which often suffer from limited capacity or linear parameter growth. \textbf{(c-d) Self-expansion paradigm (right)}: \textbf{(c)} Existing dynamic expansion methods rely solely on image perception strategies and visual shifts. \textbf{(d)} Our method incorporates brain lesion concepts to simulate the clinical cognitive process, guiding expert routing and architectural growth through stable concept reasoning.}
    \label{fig:Figure1}
\end{figure}

Recently, the method of using Pre-Trained Models (PTMs)~\cite{caron2021emerging,MedSAM} combined with Parameter-Efficient Fine-Tuning (PEFT)~\cite{hu2022lora,liu2022few,xu2026parameter} has shown great potential in CL.
By freezing PTMs to retain pre-trained knowledge and alleviate forgetting, these approaches adapt to new tasks by fine-tuning only a small set of lightweight modules. Typically, these modules are organized as specialized experts within a Mixture-of-Experts (MoE)~\cite{shazeer2017outrageously,dai2024deepseekmoe,mu2026comprehensivesurveymixtureofexpertsalgorithms} architecture to handle diverse data distributions.
Traditionally, many methods (Fig. \ref{fig:Figure1}(a))\cite{jia2022visual,zhou2022learning,Wang_2022_CVPR,Yu_2024_CVPR} utilize a fixed expert pool shared across all tasks, which inevitably constrains their potential due to a pre-defined capacity, making them ill-suited for long data streams. In contrast, other approaches (Fig. \ref{fig:Figure1}(b)) \cite{wang2022dualprompt,Smith_2023_CVPR,chen2024low} extend PTMs with strict task-specific experts for each incoming task to ensure adaptability, but this leads to linear parameter growth and hinders effective knowledge sharing.
As a more flexible alternative, recent studies have proposed dynamic expansion methods (Fig. \ref{fig:Figure1}(c)) \cite{Wang_2025_CVPR,11122658} based on an image-perception strategy. These approaches rely on image statistical shifts to determine when and where to add new experts.
While effective for general computer vision tasks, this paradigm faces limitations in brain lesion segmentation, where tasks differ not only in image appearance but also in underlying lesion pathophysiology, varying modalities, and scanner conditions. 
Consequently, addressing the heterogeneity across continual brain lesion segmentation tasks remains a critical challenge.


To address this task heterogeneity, simulating radiologists' cognitive process offers a desirable path\cite{gao2024training,szulewski2019starting}. In clinical practice, radiologists interpret complex pathologies through combinations of invariant hierarchical concepts, forming a structured reasoning chain. This chain sequentially interprets modality-specific signal patterns, locates anatomical regions of interest, and characterizes lesion morphological attributes. 
Although visual appearances fluctuate dramatically across lesions and modalities\cite{bankman2008handbook}, these underlying concepts remain stable. Specifically, fundamental tissues like white matter and cerebrospinal fluid provide a consistent anatomical context, while lesion attributes such as distinctive boundary shapes and spatial distribution patterns offer invariant descriptors across varying modalities.
Motivated by these observations, we argue for decoupling visual features into structured concepts to address the heterogeneity in continual brain lesion segmentation tasks, equipping the model with robust, scalable, and interpretable reasoning capabilities.

Building upon this perspective, we propose a Concept-Reasoning Expansion (CoRE) framework for continual brain lesion segmentation, which represents, to the best of our knowledge, the first CL paradigm endowed with clinical reasoning capabilities. To implement this, (1) we first construct a Brain Lesion Concept Library (BLC-Lib) by querying pre-trained Large Language Models (LLMs) to acquire hierarchical knowledge, serving as the reasoning foundation formulated as a concept space with textual descriptions that bridge clinical priors and visual representations. (2) We introduce a Concept Guided Calibration (CGC) module that aligns image tokens with the concept space. This module employs a dual routing mechanism that integrates concept-side semantic signals and image-side distributional cues, grounding expert selection in interpretable brain lesion attributes rather than relying solely on visual features. (3) We design a Concept-Driven Expansion (CDE) module that jointly evaluates concept-side routing confidence and image-side reconstruction deviation as a dual novelty detector for experts. This evaluation determines the necessity of architectural growth and ensures stable adaptation across long data streams.  Extensive experiments across 12 sequential tasks demonstrate that CoRE consistently outperforms existing state-of-the-art (SOTA) methods in both segmentation accuracy and knowledge retention, while exhibiting superior few-shot transferability and clinical interpretability.

\section{Related Work}

\noindent\textbf{Brain Lesion Segmentation.}
Recent advancements in brain lesion segmentation have been propelled by hybrid architectures and generative modeling. SOTA models like Swin UNETR~\cite{hatamizadeh2021swin} have set new benchmarks by modeling long-range dependencies to handle complex tumor morphologies. Simultaneously, generative frameworks such as VGDM~\cite{behnam2025vgdmvisionguideddiffusionmodel} leverage vision transformers within diffusion processes to refine lesion boundaries and quantify aleatoric uncertainty. The adaptation of medical foundation models, exemplified by GBT-SAM~\cite{dianaalbelda2025gbtsamparameterefficientdepthawaremodel}, has further enhanced zero-shot generalization across diverse clinical cohorts. These developments, often benchmarked on BraTS~\cite{adewole2023brain} using the nnU-Net v2~\cite{isensee2021nnu} or U-Mamba~\cite{ma2024u} frameworks, have significantly improved static segmentation performance. However, these works primarily operate under a stationary data distribution and do not consider the challenges of CL, such as catastrophic forgetting when adapting to non-stationary clinical environments, evolving scanner protocols, or sequential multi-site data.

\noindent\textbf{Continual Learning in Medical Imaging.}
CL in medical image addresses the issue of retaining knowledge across non-stationary data streams. Regularization-based methods, including elastic weight consolidation~\cite{kirkpatrick2017overcoming} and memory-aware synapses~\cite{aljundi2017expert}, mitigate catastrophic forgetting by penalizing updates to parameters critical for previously learned tasks. Rehearsal-based strategies utilize experience replay via memory buffers~\cite{kaustaban2022characterizing} or leverage generative replay~\cite{kumari2025domain} and latent replay~\cite{thandiackal2024multi} to address privacy concerns by synthesizing data or storing hidden representations. Finally, architecture-based techniques~\cite{bayasi2024biaspruner} isolate parameters through domain-specific batch normalization or dynamically expand the network using out-of-distribution (OOD) detection to adapt to shifting clinical data distributions. These paradigms enable models to maintain sustainability and stability across evolving healthcare environments. However, most existing methods often fail to integrate the rich, structured textual concepts inherent in medical imaging, which could otherwise provide vital semantic guidance for more reliable knowledge transfer.

\begin{figure}[t!] 
    \centering     
    \includegraphics[width=\linewidth]{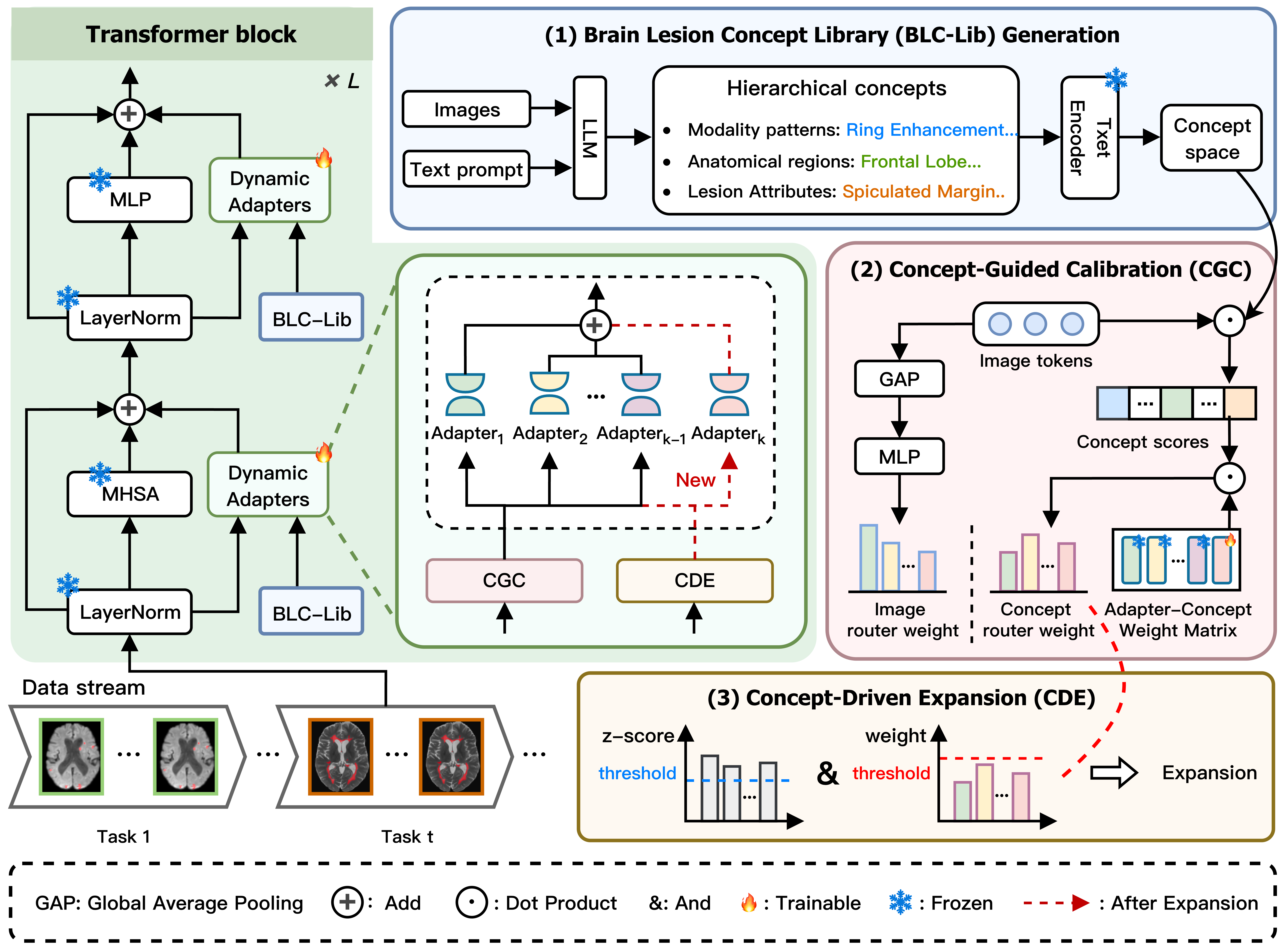} 
    \caption{Overview of the CoRE framework. The architecture consists of three main components: (\textbf{1) Brain Lesion Concept Library Generation}, which establishes a structured conceptual foundation by organizing hierarchical knowledge into a concept space; (2) \textbf{Concept-Guided Calibration (CGC)}, a module that aligns visual tokens with the concept space to ground expert routing decisions in interpretable brain lesion attributes; and (3) \textbf{Concept-Driven Expansion (CDE)}, a novelty detector that triggers dynamic architectural growth upon detecting joint visual and conceptual distribution shifts.}
    \label{fig:Figure2}
\end{figure}

\section{Method}

\subsection{Problem Definition}

We formulate the continuous adaptation of brain lesion segmentation to novel MRI modalities and disease categories as CL under domain shift without any historical data retention.
Given a dynamic data stream $\mathcal{D} = \{\mathcal{D}_1, \cdots, \mathcal{D}_T\}$, each task $\mathcal{D}_t = \{ (x_{i,t}, y_{i,t}) \}$ introduces a specific MRI modality and a novel set of lesion categories $\mathcal{C}^t$. Here, $x_{i,t} \in \mathcal{X}$ represents a 3D brain MRI volume and $y_{i,t} \in \mathcal{Y}$ denotes its corresponding segmentation mask. The segmentation model $F_\theta: \mathcal{X} \rightarrow \mathcal{Y}$ must expand its unified label space to $\mathcal{Y}^t = \bigcup_{j=1}^t \mathcal{C}^j$ using solely the current dataset $\mathcal{D}_t$ (i.e., buffer-free). During task-agnostic inference, the model predicts across the cumulative space: $y_i = \arg\max_{c \in \mathcal{Y}^t} P(y_i=c | x_{i,t})$. The goal is to optimize $\theta$ to minimize the cumulative segmentation error over $\mathcal{D}$.

\subsection{Overview}
We propose the CoRE framework to tackle the challenges of continual brain lesion segmentation under dynamic distribution shifts. Our approach is built upon the Swin UNETR architecture. This hierarchical transformer-based backbone is capable of capturing long-range dependencies and multi-scale contexts for precise segmentation. Inspired by adapter-based methods\cite{Wang_2025_CVPR, li2024atlasadapterbasedmultimodalcontinual}, we employ lightweight adapters as experts to facilitate CL. This strategy decouples the adaptation process into a low-rank trainable space, enabling efficient knowledge accumulation. The overall framework is illustrated in Fig. \ref{fig:Figure2}. Here, BrainCL serves as the fundamental knowledge base, providing structured concepts of invariant brain lesion attributes. To utilize this knowledge, CGC aligns image features with the concept space to guide expert routing decisions, while CDE functions as a novelty detector to dynamically determine required architectural growth.




\subsection{Brain Lesion Concept Library Generation}
\label{sec:braincl}
The objective of this module is to construct BLC-Lib, a comprehensive library of hierarchical concepts tailored for brain lesion segmentation. We design a two-stage knowledge elicitation pipeline leveraging an advanced large vision-language model to dynamically generate these concepts.

In the first stage, we utilize a textual retrieval prompting strategy to acquire a coarse set of theoretical brain lesion concepts. We query the LLM using specific templates designed to cover three dimensions comprising modality patterns, anatomical regions, and lesion attributes. The prompts are formulated as follows:
\begin{itemize}
    \item \textit{What are the typical signal intensity patterns of [Lesion] on [Modality]?}
    \item \textit{Where are [Lesion] lesions mainly located in the brain anatomy?}
    \item \textit{Describe the morphological characteristics (e.g., margins, texture, shape) of [Lesion] comprehensively.}
\end{itemize}
In these templates, \textit{[Lesion]} represents the specific brain lesion while \textit{[Modality]} denotes the MRI sequence type. These prompts are crafted to extract sufficient theoretical background from the internal knowledge base of the model.

Next, we refine this coarse set by leveraging the visual-question-answering capabilities of GPT-5 to align theoretical medical priors with specific data distributions. We present the model with 32 representative  cases per task, each paired with ground-truth masks, and prompt it with the following query:
\begin{itemize}
    \item \textit{Please provide visual features regarding signal intensity, anatomical location, and morphology of these masked lesion images, which contain 32 cases.}
\end{itemize}

Upon obtaining visual descriptions that characterize the prevalent attributes of these cases, we intersect them with the theoretical concepts to generate the final BLC-Lib. We denote this robust knowledge base as a concept library $C = \{c_1, c_2, ..., c_M\}$ comprising $N$ distinct brain lesion concepts. To project these textual concepts into a computable feature space, we utilize the text encoder $\mathcal{T}$ from BiomedCLIP~\cite{zhang2023biomedclip}, a pretrained medical multimodal alignment model. Specifically, we construct a concept space embedding matrix $\hat{C} \in \mathbb{R}^{M \times d_t}$, where each row of $\hat{C}$ is the extracted text 
feature $\mathcal{T}(c) \in \mathbb{R}^{d_t}$ for concept $c$.

\subsection{Concept-Guided Calibration}
\label{sec:CGC}

The CGC optimizes expert routing by fusing structured brain lesion concepts with visual feature distributions. This dual-routing mechanism ensures adapter selection is grounded in both semantic attributes and representational cues.

\noindent\textbf{Adapter Expert Modules.} 
At layer $l$ of the Swin UNETR encoder, we deploy $K^{l,s}$ adapters parallel to the MHSA ($s=\text{attn}$) and MLP ($s=\text{ffn}$) sub-layers. For input $\mathbf{x}^{l,s} \in \mathbb{R}^{N^{l,s} \times d^{l,s}}$, the $k$-th adapter applies a bottleneck transformation: $f_k^{l,s}(\mathbf{x}^{l,s}) = \mathrm{ReLU}(\mathbf{x}^{l,s}\mathbf{W}_{\mathrm{down},k}^{l,s})\mathbf{W}_{\mathrm{up},k}^{l,s}$, where $\mathbf{W}_{\mathrm{down},k}^{l,s}$ and $\mathbf{W}_{\mathrm{up},k}^{l,s}$ are projection matrices with bottleneck rank $r \ll d^{l,s}$.

\noindent\textbf{Concept-Side Routing.} 
We first project the concept matrix $\hat{\mathbf{C}} \in \mathbb{R}^{M \times d_t}$ into the visual space via a learnable mapping $\phi^{l,s}$, yielding $\tilde{\mathbf{C}}^{l,s} \in \mathbb{R}^{M \times d^{l,s}}$. Token-concept similarities are computed as $\mathbf{S}^{l,s} = \mathbf{x}^{l,s} (\tilde{\mathbf{C}}^{l,s})^{\top}$ and mean-pooled into an image-level activation $\bar{\mathbf{s}}^{l,s} \in \mathbb{R}^{M}$. To associate concepts with adapters, we maintain a adapter-concept weight matrix $\mathbf{W}_\text{AC}^{(l,s)} \in \mathbb{R}^{M \times K^{l,s}}$, where individual columns serve as semantic descriptors for the available adapters while rows represent the pre-defined brain lesion concepts. The concept routing weight is then $\mathbf{w}^{l,s}_{\mathrm{c}} = \mathrm{Softmax}(\bar{\mathbf{s}}^{l,s}\mathbf{W}_\text{AC}^{(l,s)})$. To prevent catastrophic forgetting, only the column corresponding to the newly added adapter is updated, while existing columns are kept fixed.

\noindent\textbf{Image-Side Routing.} 
Parallelly, an image-side router captures distributional characteristics omitted by discrete concepts. Using spatially pooled features $\bar{\mathbf{x}}^{l,s} \in \mathbb{R}^{d^{l,s}}$, the visual weight is $\mathbf{w}^{l,s}_{\mathrm{v}} = \mathrm{Softmax}(\bar{\mathbf{x}}^{l,s}\mathbf{W}^{l,s}_{\mathrm{r}})$, where $\mathbf{W}^{l,s}_{\mathrm{r}}$ is the expandable router matrix. Similar to the concept side, only newly appended columns are trainable.

\noindent\textbf{Multimodal Routing Fusion.} 
The final routing weight $\mathbf{w}^{l,s} = \lambda \mathbf{w}^{l,s}_{\mathrm{c}} + (1-\lambda) \mathbf{w}^{l,s}_{\mathrm{v}}$ fuses both signals. The adapter outputs are then weighted and aggregated with the standard sub-layer computation $\Phi^{l,s}(\mathbf{x}^{l,s})$:
\begin{equation}
    \mathbf{x}^{l,s}_{\mathrm{out}} = \Phi^{l,s}(\mathbf{x}^{l,s}) + \sum_{k=1}^{K^{l,s}} w_k^{l,s} \cdot f_k^{l,s}(\mathbf{x}^{l,s})
\end{equation}
This fusion equips each transformer block with adaptive routing suited for heterogeneous brain lesion data.

\subsection{Concept-Driven Expansion}
\label{sec:CDE}

The CDE module governs dynamic architectural growth. During the initial training epoch of a new task $\mathcal{D}_t$, CDE evaluates whether existing adapters can handle the incoming data based on concept-side and image-side signals. A new adapter is instantiated only if both confirm distributional novelty.

\noindent\textbf{Concept-Side Expansion Signal.} The routing weight $\mathbf{w}_{\mathrm{c}}^{l,s}$ from the CGC module reflects conceptual familiarity. A flat distribution implies no single existing adapter adequately covers the input's conceptual identity. We average these weights over the $n_t$ samples of the new task: $\bar{\mathbf{w}}_{\mathrm{c}}^{l,s} = \frac{1}{n_t} \sum_{i=1}^{n_t} \mathbf{w}_{\mathrm{c},i}^{l,s}$. The concept-side signal is triggered if the maximum activation $\max(\bar{\mathbf{w}}_{\mathrm{c}}^{l,s}) < \tau_{\mathrm{c}}$, where $\tau_{\mathrm{c}} \in (0,1)$ is a confidence threshold, indicating the presence of novel conceptual patterns.

\noindent\textbf{Image-Side Expansion Signal.} To capture local feature distributions, each adapter is paired with a lightweight autoencoder $d_k^{l,s}$. It is trained to reconstruct the specific features its adapter handles, minimizing $\mathcal{L}_{\mathrm{est},k}^{l,s} = \sum \|\mathbf{x}^{l,s} - d_k^{l,s}(\mathbf{x}^{l,s})\|_2^2$. For incoming samples, we compute the z-score of the reconstruction error, $\zeta_k^{l,s}$, normalized against running statistics to remove scale variations across tasks. The image-side signal is triggered if the average z-score across all new samples exceeds a threshold $\tau_{\mathrm{i}}$ for every existing estimator at position $(l,s)$.

\noindent\textbf{Joint Expansion Decision.} A new adapter is introduced only when both signals activate, preventing redundant expansion. Once trained on $\mathcal{D}_t$, the new parameters are locked to mitigate catastrophic forgetting. If no expansion occurs, existing modules are seamlessly reused without parameter updates.

\noindent\textbf{Training Objective.} The overall objective jointly optimizes the segmentation model $F_\theta$ and the estimators $d_k^{l,s}$. We employ a composite segmentation loss $\mathcal{L}_\text{seg} = 0.8\mathcal{L}_\text{Dice} + 0.2\mathcal{L}_\text{BCE}$, where $\mathcal{L}_\text{Dice}$ represents Dice loss and $\mathcal{L}_\text{BCE}$ represents Binary Cross-Entropy (BCE) loss. The complete training objective across the task sequence is:
\begin{equation}
    \min_{\theta,\,\{d_k^{l,s}\}}
    \sum_{t=1}^{T} \left[
    \mathbb{E}_{(\mathbf{x},\mathbf{y})\in\mathcal{D}_t}  \mathcal{L}_{\text{seg}}\!\left(F_\theta(\mathbf{x}),\,\mathbf{y}\right)
    + \sum_{l,s}\sum_{k=1}^{K^{l,s}}  \mathcal{L}_{\mathrm{est},k}^{l,s}\!\left(\mathbf{x}^{l,s}\right)
    \right].
\end{equation}
If no expansion occurs, existing modules are seamlessly reused without parameter updates.

\begin{table}[t]
\centering
\caption{Comparison with the SOTA methods on brain lesion segmentation by consecutive learning on the data stream from Task 1 to 12. Bold font represents the best result (excluding offline JointTrain and task-specific IndividualTrain).}
\label{tab:comprehensive_results}
\begin{small}
\resizebox{\textwidth}{!}{%
\begin{tabular}{l cccc c cccc c cc c c}
\toprule[1.3pt]

\multirow{5}{*}{\textbf{Method}} 
& \multicolumn{4}{c}{BraTS} 
& ATLAS 
& \multicolumn{4}{c}{MSSEG} 
& ISLES & \multicolumn{2}{c}{WMH} 
& \multirow{5}{*}{\textbf{Average}} 
& \multirow{5}{*}{\textbf{BWT}} \\
\cmidrule(lr){2-5} \cmidrule(lr){6-6} \cmidrule(lr){7-10} \cmidrule(lr){11-11} \cmidrule(lr){12-13}

& T1w & T2w & T1ce & FLAIR & T1w & T1w & T2w & T1ce & FLAIR & DWI & T1w & FLAIR & & \\
\cmidrule(lr){2-5} \cmidrule(lr){6-6} \cmidrule(lr){7-10} \cmidrule(lr){11-11} \cmidrule(lr){12-13}

& Task 1 & Task 2 & Task 3 & Task 4 & Task 5 & Task 6 & Task 7 & Task 8 & Task 9 & Task 10 & Task 11 & Task 12 & & \\
\cmidrule(lr){2-13}

\multicolumn{15}{c}{\textbf{Dice Similarity Coefficient (DSC) (\%) $\bm{\uparrow}$}} \\
\midrule
\rowcolor{headergray} IndividualTrain & 85.78 & 90.35 & 86.35 & 92.81 & 71.28 & 70.81 & 72.49 & 70.50 & 80.91 & 87.98 & 76.77 & 83.62 & 80.80 & 100.00 \\
JointTrain & 84.78 & 90.18 & 87.02 & 92.67 & 70.31 & 70.62 & 73.63 & 70.18 & 79.82 & 86.01 & 75.23 & 81.98 & 80.20 & 99.33\\
Finetune & 26.59 & 73.11 & 38.86 & 86.46 & 3.57 & 8.27 & 23.67 & 0.19 & 70.90 & 80.99 & 16.92 & 83.60 & 42.76 & 58.50\\
\midrule

L2P & 73.56 & 85.65 & 78.36 & 90.16 & 59.66 & 57.96 & 60.56 & 45.12 & 75.27 & 82.67 & 71.06 & 80.67 & 71.73 & 90.36\\

Moe-adapters & 73.78 & 85.32 & 78.57 & 90.09 & 60.91 & 60.45 & 61.75 & 45.98 & 76.23 & 82.22 & 72.63 & 81.26 & 72.43 & 91.08\\

DualPrompt & 74.62 & 85.78 & 80.14 & 90.25 & 61.99 & 62.09 & 62.12 & 46.79 & 76.02 & 83.18 & 70.84 & 83.26 & 73.16 & 91.69\\

CODA-P & 76.79 & 86.11 & 80.48 & 90.69 & 62.70 & 62.24 & 61.91 & 46.07 & 78.36 & 83.51 & 73.39 & \textbf{83.87} & 73.84 & 92.38\\

Low-Rank MoE & 78.23 & 86.37 & 81.22 & 90.10 & 62.23 & 65.19 & 66.78 & 58.23 & 78.22 & 84.12 & 72.26 & 83.12 & 75.51 & 94.27\\

SEMA & 82.86 & 87.56 & 84.38 & 90.18 & 65.02 & 65.39 & 68.34 & 61.97 & 79.93 & 85.71 & 74.53 & 83.17 & 77.42 & 96.35\\

Moe-adapters++ & 81.52 & 87.57 & 84.52 & 90.98 & 64.19 & 68.42 & 65.25 & 63.66 & \textbf{80.78} & 84.90 & 74.01 & 83.43 & 77.44 & 96.34\\

\rowcolor{color3} \textbf{CoRE (ours)} & \textbf{85.64} & \textbf{90.11} & \textbf{86.89} & \textbf{92.87} & \textbf{68.15} & \textbf{70.83} & \textbf{70.41} & \textbf{66.37} & 79.86 & \textbf{87.54} & \textbf{76.51} & 83.66 & \textbf{79.90} & 98.96\\
\bottomrule[1.3pt]

\end{tabular}}
\end{small}
\end{table}

\section{Experiments}
\subsection{Experiments Setup}
\noindent\textbf{Datasets.}
We evaluate CoRE on five publicly available multi-modal MRI datasets for brain lesion segmentation: 5,004 subjects in BraTS\cite{baid2021rsna} (Tasks 1 to 4, T1w\slash T2w\slash T1ce\slash FLAIR), 655 subjects in ATLAS\cite{liew2018large} (Task~5, T1w), 60 subjects in MSSEG\cite{commowick2018objective}  (Tasks~6--9, T1w/T2w/T1ce/FLAIR), 250 subjects in ISLES\cite{hernandez2022isles} (Task~10, DWI), and 120 subjects in WMH\cite{kuijf2019standardized} (Tasks~11--12, T1w/FLAIR). The model is trained sequentially from Task~1 to Task~12. To further evaluate CoRE under few-shot multi-domain class incremental learning, 240 subjects in BraTS2023-SSA\cite{adewole2023brain} (Tasks~13--16, T1w/T2w/T1ce/FLAIR) are additionally introduced as a downstream evaluation dataset. For pre-processing, all images were rigidly registered to the MNI152 template using ANTs\cite{avants2011reproducible}. Brain masks were applied for skull removal, and all volumes were cropped to a standardized size of $160\times196\times160$ with a voxel spacing of $1\times1\times1$~mm$^3$. Within each task, subjects were divided into training, validation, and test sets at a ratio of 70\%, 10\%, and 20\%, respectively. More details are provided in the Appendix.

\noindent\textbf{Metrics.}
Segmentation accuracy is evaluated using the Dice Similarity Coefficient (DSC) on each task. To comprehensively assess knowledge retention, we additionally report Backward Transfer (BWT), which measures the average performance change on earlier tasks after learning subsequent ones. An advanced continual learning method should achieve high DSC alongside high BWT.

\noindent\textbf{Implementation details.}
We employ the improved Swin UNETR from MONAI as our backbone, initialized with pre-trained weights from 2,100 CT volumes\cite{li2024well}. To bridge the substantial domain gap between CT pre-training and MRI-based tasks, only the decoder is finetuned on Task~1, while the entire backbone remains frozen for all subsequent tasks. All experiments are conducted on two NVIDIA GeForce RTX~4090 GPUs. We train the model for 300 epochs per task using the AdamW optimizer with a batch size of~8. The learning rate is initialized at $3 \times 10^{-4}$ with a weight decay of $1 \times 10^{-4}$, following a cosine annealing schedule with a 20-epoch warm-up. For our proposed modules, the balance hyperparameter $\lambda$ in CGC is set to 0.7. In CDE, the concept-side threshold $\tau_{\mathrm{c}}$ and image-side threshold $\tau_{\mathrm{i}}$ are empirically configured to 0.7 and 1.3, respectively.

\begin{figure}[t!] 
    \centering     
    \includegraphics[width=\linewidth]{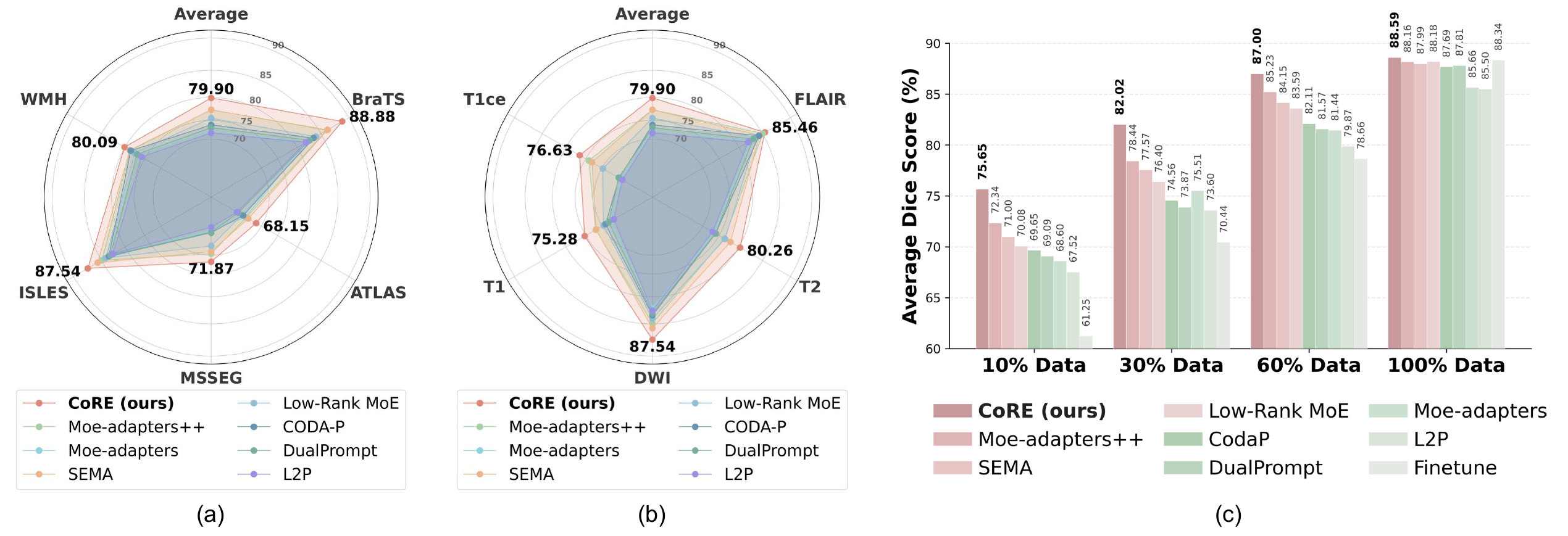} 
    \caption{Quantitative performance and data efficiency analysis. \textbf{(a)} Lesion-level performance, comparing the average DSC of CoRE against comparative methods across diverse brain lesion types; \textbf{(b)} modality-level performance, comparing the average DSC against comparative methods across various imaging modalities; and \textbf{(c)} few-shot multi-domain class incremental learning, illustrating the average DSC for Tasks 13–16 under varying training data volumes (10\%, 30\%, 60\%, and 100\%).}
    \label{fig:radar-bar}
\end{figure}

\subsection{Comparison with State-of-the-art Methods}
In this section, we evaluate the performance of CoRE against several PTM-based rehearsal-free CL approaches using the same backbone. These baselines are categorized into: (1) fixed expert pool method: L2P~\cite{Wang_2022_CVPR} and Moe-adapters \cite{Yu_2024_CVPR}; (2) task-specific expansion method: DualPrompt \cite{wang2022dualprompt}, CODA-P \cite{Smith_2023_CVPR}, and Low-Rank MoE \cite{chen2024low}; and (3) image-perception expansion method: SEMA \cite{Wang_2025_CVPR} and Moe-adapters++ \cite{11122658}.

\noindent\textbf{Preliminary Experiments and Analysis.}
To investigate whether and how the model suffers from forgetting and domain shifts, we initially set up three baseline scenarios: \textbf{IndividualTrain}, which maintains separate models for each task; \textbf{JointTrain}, representing the performance upper bound through offline multi-task learning; and \textbf{Finetune}, which involves sequential adaptation without forgetting mitigation. 

As observed in Table \ref{tab:comprehensive_results}, IndividualTrain achieves optimal performance across all tasks with zero forgetting (100\% BWT). However, this approach requires additional storage capacity for maintaining incremental models, making it both inefficient and inflexible for real-world clinical deployment. In contrast, our method achieves comparable performance using a single model. The performance of Finetune exhibits a significant performance drop on earlier tasks, achieving an average DSC of only 42.76\% and a BWT of 58.50\%. These observations demonstrate that task-by-task finetuning leads to catastrophic forgetting, as the model overfits to incoming task distributions at the cost of previously learned knowledge.

\begin{figure}[t!] 
    \centering     
    \includegraphics[width=\linewidth]{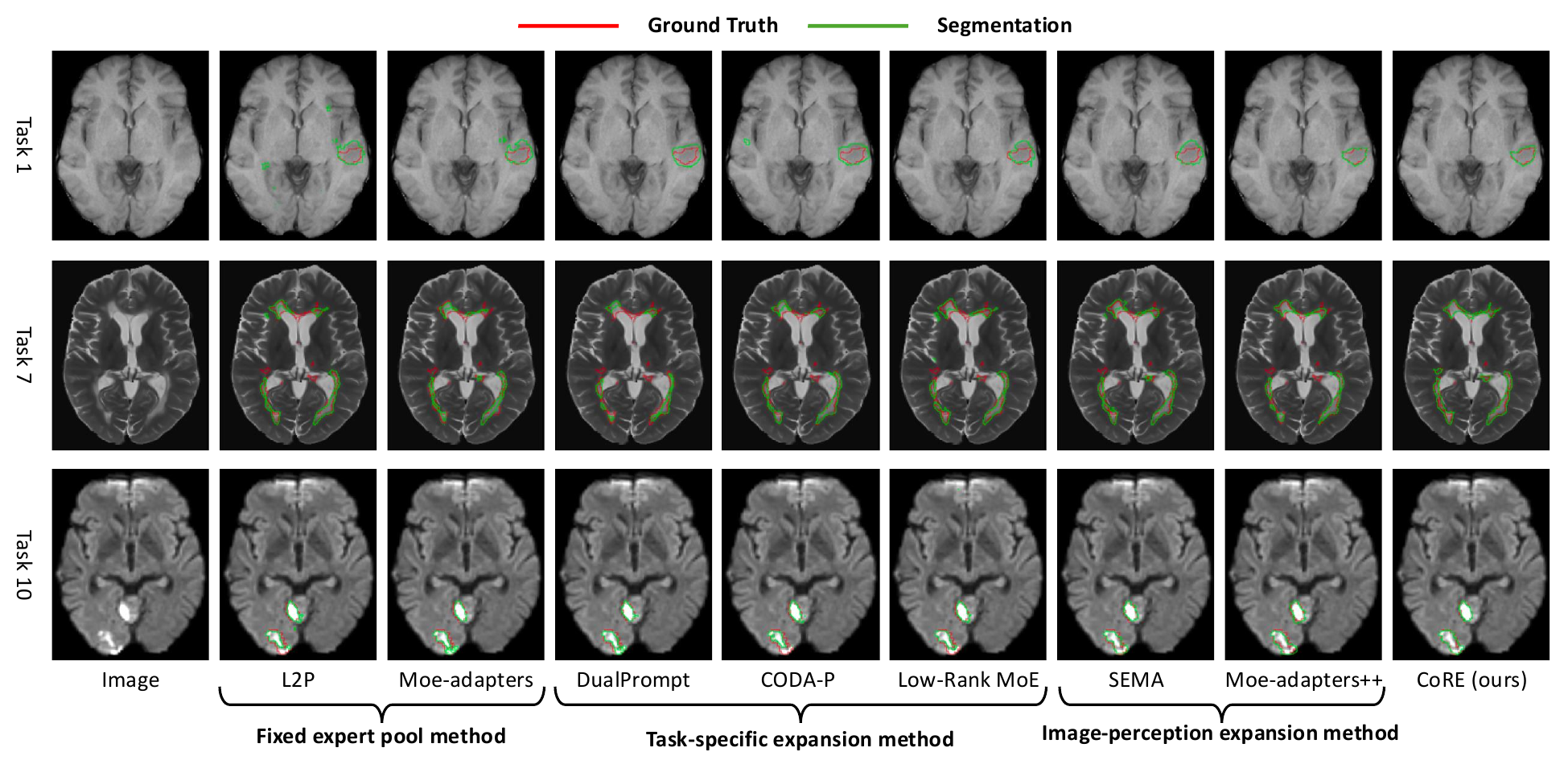} 
    \caption{Qualitative comparison of brain lesion segmentation results in the data stream from Task 1 to 12.}
    \label{fig:visualization}
\end{figure}

\noindent\textbf{Multi-domain Class Incremental Learning.}
The comparative results for the 12-task data stream are presented in Table \ref{tab:comprehensive_results}. Generally, dynamic expansion methods outperform fixed-pool and task-specific approaches, which aligns with the architectural advantages of self-expansion mechanisms in handling long-term data streams. Fixed-pool methods struggle to accommodate these streams due to their pre-defined capacity, resulting in limited accuracy on Task~12. While task-specific extension methods are highly adaptable, they typically exhibit low BWT due to limited knowledge reuse across different tasks, leading to a high forgetting rate.
CoRE successfully optimizes the model for both stability and plasticity. It outperforms all compared methods, achieving a 2.46\% improvement in overall DSC and a 2.61\% increase in BWT over the second-best model. Furthermore, as illustrated in the radar charts (Fig. \ref{fig:radar-bar} (a) and (b)), CoRE achieves the highest performance across diverse lesion types and imaging modalities, highlighting its powerful capability to handle complex brain lesion segmentation tasks. We provide a qualitative comparison in Fig. \ref{fig:visualization}, where our proposed method demonstrates the closest approximation to the ground truth among all compared approaches.

\noindent\textbf{Few-shot Multi-Domain Class Incremental Learning.}
A robust CL model should not only retain past knowledge but also leverage accumulated insights to facilitate faster and more efficient learning of future tasks. We evaluate this capability on the BraTS2023-SSA dataset using 10\%, 30\%, 60\%, and 100\% of the training samples, reporting the average DSC over 5 runs to reduce variability. Fig.~\ref{fig:radar-bar} (c) illustrates the average experimental results for Tasks 13--16 across different data volumes. As shown, Finetune exhibits poor few-shot performance (61.25\% DSC at 10\% data), indicating that sequential updates struggle to learn effective representations from limited samples. While other expansion methods maintain competitive performance with complete data, their accuracy drops sharply under limited data conditions. In contrast, CoRE leverages its accumulated conceptual knowledge for rapid task adaptation, enabling high segmentation accuracy even in data-constrained scenarios and demonstrating superior transferability. Detailed results for Tasks 13--16 are provided in the Appendix.

\subsection{Extensive Analysis}
\label{sec:extensive_analysis}

\begin{figure}[t!] 
    \centering     
    \includegraphics[width=\linewidth]{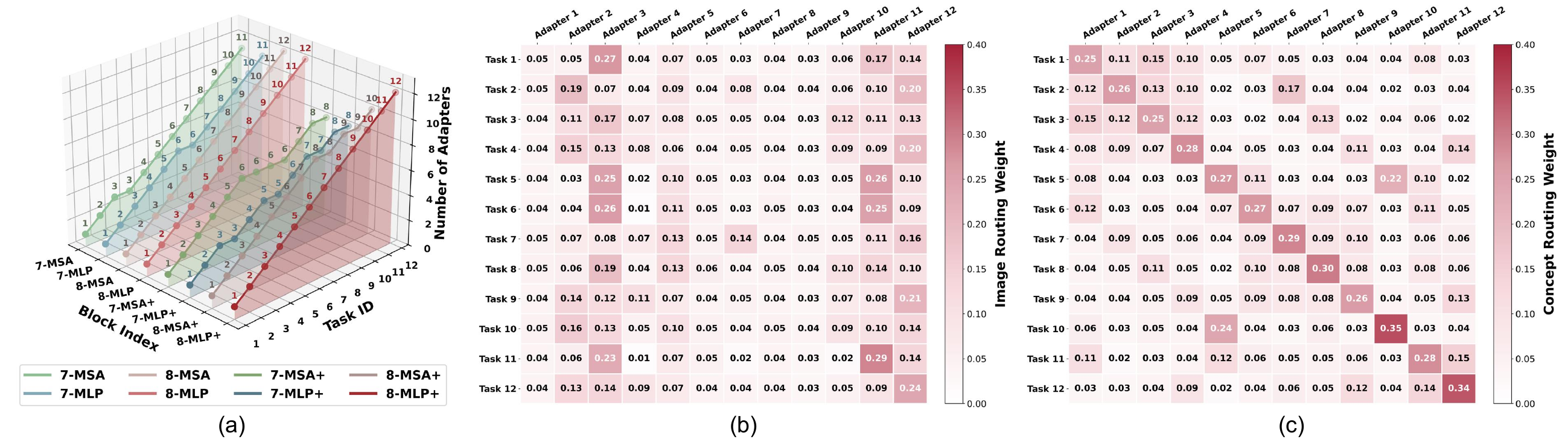} 
    \caption{Analysis of dynamic adapter expansion and routing weight distributions. \textbf{(a)} Dynamic adapter expansion, illustrating the number of adapters across the 12-task data stream for Block 7 and Block 8, where ‘+’ denotes layers utilizing concept-driven expansion; \textbf{(b)} Image routing weights and \textbf{(c)} concept routing weights, showing the weight distributions of the 8-MLP layer across the 12 sequential tasks.}
    \label{fig:heatmap-waterfall}
\end{figure}

\noindent\textbf{Analysis of Dynamic Adapter Expansion.}
Guided by preliminary experiments, we restrict dynamic expansion to the final two transformer blocks, specifically Block 7 and Block 8. We found that allowing expansion in shallow layers increases the total adapter count without notable performance gains, likely because these layers primarily capture low-level visual textures that behave similarly despite distribution shifts across MRI modalities. Fig. \ref{fig:heatmap-waterfall} (a) illustrates adapter growth in the expandable layers across 12 tasks. Block indices marked with a `+' denote our concept-driven expansion, whereas those without rely solely on image features. As observed, purely image-based expansion triggers nearly linear growth, indicating extreme sensitivity to lesion and modality shifts, thereby causing parameter redundancy. Conversely, integrating concepts effectively regulates this growth by conditioning expansion decisions on conceptual novelty. Consequently, 7-MSA+ and 7-MLP+ stabilize at only 8 adapters, while 8-MSA+ reaches 10. By achieving optimal segmentation performance with a more compact architecture, CoRE demonstrates that concept guidance successfully identifies task similarities, mitigating unnecessary capacity increases and maximizing knowledge reuse.

\noindent\textbf{Analysis of Routing Weight Distributions.}
To investigate the routing behavior, we visualize the weight distributions of the 8-MLP layer across the 12 sequential tasks. As illustrated in Fig. \ref{fig:heatmap-waterfall} (b), regardless of the input task, the image routing weights consistently concentrate on a few specific modules, such as Adapters 3, 11, and 12.
Such behavior suggests an inherent difficulty in disentangling diverse brain diseases or modalities using solely low-level visual features, ultimately leading to an over-reliance on a small subset of adapters. However, the image-side routing mechanism remains crucial, as it is specifically designed to capture continuous distribution characteristics that discrete concepts cannot fully express, providing essential visual perception. Fig. \ref{fig:heatmap-waterfall} (c) displays the concept routing weights, which exhibit a clear diagonal trend. With the introduction of BLC-Lib, CoRE successfully decouples complex visual features into structured medical knowledge, explicitly assigning each adapter to process specific lesion concepts. Furthermore, observing off-diagonal highlights reveals further insights. For example, the strong activation of Adapter 10 in Task 5 suggests that concept routing is highly sensitive to overlapping brain lesion concepts (e.g., shared disease categories) across different tasks, thus enabling the intelligent reuse of learned adapters. Ultimately, the concept-side routing provides a discrete, structured medical prior boundary that complements the continuous image-side routing to ensure accurate expert selection.
\begin{figure}[t!] 
    \centering     
    \includegraphics[width=\linewidth]{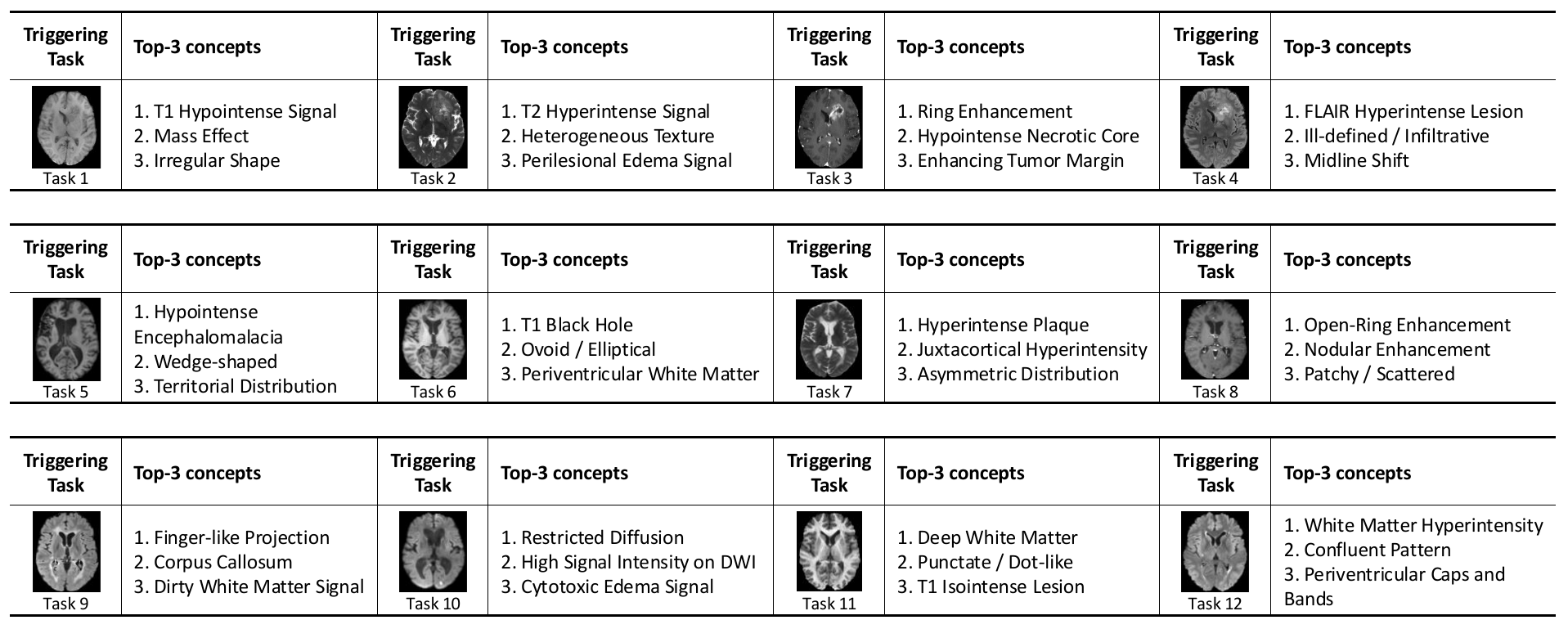} 
    \caption{Concept-adapter affinity analysis of the 8-MLP layer. For each of the 12 tasks, we display the top-3 brain lesion concepts associated with the adapter introduced at that task, extracted from the highest-weighted entries in the corresponding column of the adapter-concept weight matrix $\mathbf{W}_\text{AC}$.}
    \label{fig:top-3 concept}
\end{figure}

\noindent\textbf{Analysis of Concept-Adapter Affinity.}
To demonstrate CoRE’s interpretability, we visualize the top-3 brain lesion concepts per adapter in the 8-MLP layer (Fig.~\ref{fig:top-3 concept}). This layer allocates one adapter per task, yielding 12 total. 
These representative concepts are extracted from the highest-weighted rows in the $\mathbf{W}_\text{AC}$. The results reveal that each adapter converges upon a clinically coherent conceptual profile. For example, the first four adapters triggered by BraTS tasks successfully disentangle tumor attributes driven by different modalities, capturing different facets of the same underlying pathology. This specialization within lesions elucidates the dense local mutual activations observed across Tasks 1 to 4 in Fig. ~\ref{fig:heatmap-waterfall} (c). By establishing precise conceptual boundaries, CGC prevents interference across tasks even within identical disease categories. A similar pattern is additionally observed in the MSSEG group across Tasks 6 to 9.
Beyond specialization within lesions, the affinity also reveals meaningful complementarity across tasks. For instance, Tasks 5 and 10 both represent stroke lesions. Although they encode distinct chronic and acute markers, respectively, their shared ischemic pathophysiology induces mutual activation across these tasks.

Taken together, these observations confirm that BLC-Lib provides a structured prior that enables the model to ground its routing and expansion decisions in interpretable brain lesion concepts. Furthermore, the $\mathbf{W}_\text{AC}$ explicitly specializes each adapter in a distinct conceptual profile, supporting both discriminative knowledge partitioning and principled cross-task knowledge sharing aligned with radiological reasoning.

\begin{table}[t]
\centering
\caption{Ablation study of each component in our proposed approach.} 
\label{tab:ablation_results}
\begin{small}
\resizebox{\textwidth}{!}{%
\begin{tabular}{l cccc c cccc c cc c c}
\toprule[1.3pt]

\multirow{5}{*}{\textbf{Method}} 
& \multicolumn{4}{c}{BraTS} 
& ATLAS 
& \multicolumn{4}{c}{MSSEG} 
& ISLES & \multicolumn{2}{c}{WMH} 
& \multirow{5}{*}{\textbf{Average}} 
& \multirow{5}{*}{\textbf{BWT}} \\
\cmidrule(lr){2-5} \cmidrule(lr){6-6} \cmidrule(lr){7-10} \cmidrule(lr){11-11} \cmidrule(lr){12-13}

& T1w & T2w & T1ce & FLAIR & T1w & T1w & T2w & T1ce & FLAIR & DWI & T1w & FLAIR & & \\
\cmidrule(lr){2-5} \cmidrule(lr){6-6} \cmidrule(lr){7-10} \cmidrule(lr){11-11} \cmidrule(lr){12-13}

& Task 1 & Task 2 & Task 3 & Task 4 & Task 5 & Task 6 & Task 7 & Task 8 & Task 9 & Task 10 & Task 11 & Task 12 & & \\
\cmidrule(lr){2-13}

\multicolumn{15}{c}{\textbf{Dice Similarity Coefficient (DSC) (\%) $\bm{\uparrow}$}} \\
\midrule

Finetune & 26.59 & 73.11 & 38.86 & 86.46 & 3.57 & 8.27 & 23.67 & 0.19 & 70.90 & 80.99 & 16.92 & 83.60 & 42.76 & 58.50\\
Raw.C & 84.76 & 88.87 & 84.77 & 90.34 & 66.52 & 68.76 & 68.11 & 65.36 & 78.47 & 84.37 & 72.71 & 82.69 & 77.98 & 97.00 \\
Rand.C & 80.33 & 85.46 & 80.86 & 87.11 & 61.14 & 63.16 & 63.59 & 60.04 & 75.00 & 80.12 & 68.11 & 79.08 & 73.67 & 92.63 \\
CoRE (w/o CGC) & 83.47 & 88.15 & 84.52 & 90.92 & 65.11 & 68.05 & 68.24 & 64.02 & 78.60 & 84.15 & 74.37 & 82.09 & 77.64 & 96.69 \\
CoRE (w/o CDE) & 82.37 & 88.79 & 84.67 & 90.14 & 65.90 & 68.46 & 68.25 & 65.88 & 78.92 & 86.82 & 75.09 & 83.12 & 78.20 & 97.21  \\
\rowcolor{color3}\textbf{CoRE (ours)} & \textbf{85.64} & \textbf{90.11} & \textbf{86.89} & \textbf{92.87} & \textbf{68.15} & \textbf{70.83} & \textbf{70.41} &\textbf{ 66.37} & \textbf{79.86 }& \textbf{87.54} & \textbf{76.51} & \textbf{83.66} & \textbf{79.90 }& \textbf{98.96} \\
\bottomrule[1.3pt]

\end{tabular}%
}
\end{small}
\end{table}

\subsection{Ablation Study}
We conduct systematic ablation experiments to validate the contribution of each proposed component, with results summarized in Table. \ref{tab:ablation_results}.

\noindent\textbf{Effectiveness of BLC-Lib Construction}. To assess the quality of our two-stage concept elicitation pipeline, we compare it against two degraded variants. Raw.C replaces the two-stage pipeline with a single-round LLM query, yielding 77.98\% DSC and 97.00\% BWT. Although this simplified strategy still provides semantic guidance, the resulting concepts lack sufficient grounding in task-specific visual observations, leading to suboptimal routing and expansion decisions. Rand.C further replaces the concept library with randomly generated, task-irrelevant descriptions, causing a pronounced drop to 73.67\% DSC and 92.63\% BWT. This degradation confirms that the performance gains of CoRE are not merely attributable to the concept-routing mechanism itself, but critically depend on the clinical relevance and structural precision of the underlying knowledge base. These results collectively validate that the two-stage elicitation process is essential for constructing a concept library that is both medically grounded and data-aligned.

\noindent\textbf{Effectiveness of the CGC Module}. Removing the CGC module leads to a reduction in average DSC to 77.64\% and BWT to 96.69\%. Without concept-side routing, the model relies solely on visual features to select adapters, which struggles to disentangle the heterogeneous lesion distributions across tasks and modalities. As analyzed in Section~\ref{sec:extensive_analysis}, this visual-only routing tends to collapse onto a small number of dominant adapters, impairing knowledge reuse and introducing inter-task interference. The CGC module addresses this limitation by calibrating adapter selection with structured brain lesion concepts, enabling more discriminative and balanced expert utilization across the data stream.

\noindent\textbf{Effectiveness of the CDE Module}. Purely visually driven expansion without CDE triggers nearly linear adapter growth. The resulting parameter redundancy causes a notable performance drop on earlier tasks, with Task 1 dropping from 85.64\% to 82.37\%. This degradation indicates that excessive expansion creates overlapping experts, which introduces severe routing ambiguity. Consequently, the routing mechanism struggles to reliably allocate previously learned data to the correct original experts, thereby exacerbating catastrophic forgetting. By enforcing a joint visual and conceptual expansion criterion, CDE effectively suppresses this redundant growth, ultimately improving BWT by 1.75\% compared to the variant without CDE.

Overall, the full CoRE framework achieves the best performance with 79.90\% DSC and 98.96\% BWT, confirming that all three components function synergistically and are each indispensable to the final framework.

\section{Conclusion}
In this work, we presented CoRE, a concept-reasoning expansion framework that redefines the continual learning paradigm for brain lesion segmentation by shifting from image-perception strategies toward structured clinical reasoning. Rather than relying on visual statistics alone, CoRE grounds model evolution in invariant hierarchical concepts, enabling robust adaptation across heterogeneous MRI modalities and lesion categories without accessing historical data. 
Through three synergistic components (BLC-Lib, CGC, and CDE), CoRE establishes a joint decision-making mechanism that simultaneously improves expert routing precision and suppresses redundant architectural growth. Extensive experiments across 12 sequential tasks demonstrate that CoRE consistently outperforms existing SOTA methods in both segmentation accuracy and knowledge retention, while exhibiting superior few-shot transferability. 
Beyond its empirical contributions, CoRE suggests a broader principle: embedding domain-specific conceptual priors into continual learning architectures offers a principled path toward clinically trustworthy AI. Nevertheless, the current framework relies on a pre-defined concept library, which may not fully capture rare pathologies. Future work could explore adaptive concept discovery and extension to broader neurological conditions, further advancing universal clinical segmentation models.

\newpage

\bibliographystyle{splncs04}
\bibliography{main}

\newpage
\appendix

\section{Appendix Introduction}
This supplementary material provides additional details and extended analyses to support the main text. The document is organized as follows:
\begin{itemize}
    \item \textbf{Sec.~\ref{datasets}} provides comprehensive descriptions of the six datasets utilized in our experiments.
    \item \textbf{Sec.~\ref{results}} presents detailed quantitative results for few-shot multi-domain class incremental learning.
    \item \textbf{Sec.~\ref{analysis}} offers further extensive analysis, including investigations into task order sensitivity (Sec.~\ref{Task-Order}), the configuration of expandable layers (Sec.~\ref{Expandable-Layers}), the influence of the balance hyperparameter $\lambda$ (Sec.~\ref{Balance-Hyperparameter}), and the impact of different LLM generators on concept generation (Sec.~\ref{LLM-Generators}).
\end{itemize}

\section{Detailed Dataset Descriptions}
\label{datasets}








In this section, we provide comprehensive information about our experimental datasets, which introduce diverse modality and disease shifts to evaluate our continual learning framework.

\noindent\textbf{BraTS.} The BraTS dataset annotates brain gliomas across four distinct MRI modalities, specifically T1w, T2w, T1ce, and FLAIR. To evaluate continual learning under modality shifts, each imaging sequence is treated as an independent task. While the original dataset delineates the enhancing tumor, the necrotic core, and the edematous tissue, we combined these three target regions into a single whole tumor mask for our experiments.

\noindent\textbf{ATLAS.} The ATLAS dataset provides T1w MRI scans specifically focused on chronic stroke lesions. It contains manually segmented masks to accurately identify the location and extent of the infarcted tissue in the chronic stage. In our continual learning framework, this dataset introduces a significant shift in the underlying disease category.

\noindent\textbf{MSSEG.} The MSSEG dataset focuses on the segmentation of multiple sclerosis lesions across four distinct MRI modalities, specifically T1w, T2w, T1ce, and FLAIR. Consistent with our continual learning configuration, each imaging sequence is treated as an independent task.

\noindent\textbf{ISLES.} The ISLES dataset focuses on the segmentation of acute ischemic stroke lesions. Unlike the chronic stroke lesions in the ATLAS dataset, it exclusively utilizes the DWI modality to capture the acute pathophysiological characteristics of the stroke. In our continual learning framework, this dataset introduces a specific modality and disease shift to evaluate the adaptability of the model.

\noindent\textbf{WMH.} The WMH dataset focuses on the segmentation of white matter hyperintensities across two distinct MRI modalities, specifically T1w and FLAIR. Consistent with our continual learning configuration, each imaging sequence is treated as an independent task.

\noindent\textbf{BraTS2023-SSA.} The BraTS2023-SSA dataset focuses on brain glioma segmentation for Sub Saharan African populations across T1w, T2w, T1ce, and FLAIR modalities. To evaluate continual learning under modality shifts, each imaging sequence is treated as an independent task, and the original three target regions are combined into a single whole tumor mask. Unlike the BraTS dataset, these images often suffer from poor contrast and resolution due to low quality MRI technology. Additionally, the gliomas in this demographic are typically detected at advanced stages and exhibit unique characteristics like high gliosis rates. These clinical and technical variations introduce a substantial domain shift. Therefore, this dataset serves as a challenging downstream evaluation for our few shot multi domain class incremental learning framework.

\section{Detailed Results for Few-Shot Multi-Domain Class Incremental Learning}
\label{results}

\begin{table}[t]
\centering
\caption{Detailed performance comparison for few-shot multi-domain class incremental learning on the BraTS2023-SSA dataset across different training data ratios.} 
\label{tab:few_shot_results}
\begin{small}
\resizebox{\textwidth}{!}{%
\begin{tabular}{l cccc cccc cccc cccc }
\toprule[1.3pt]

\multirow{5}{*}{\textbf{Method}} 
& \multicolumn{4}{c}{Task 13} 
& \multicolumn{4}{c}{Task 14} 
& \multicolumn{4}{c}{Task 15} 
& \multicolumn{4}{c}{Task 16} 
\\
\cmidrule(lr){2-5} \cmidrule(lr){6-9} \cmidrule(lr){10-13} \cmidrule(lr){14-17} 

& \multicolumn{4}{c}{BraTS2023-SSA (T1w)} 
& \multicolumn{4}{c}{BraTS2023-SSA (T2w)} 
& \multicolumn{4}{c}{BraTS2023-SSA (T1ce)} 
& \multicolumn{4}{c}{BraTS2023-SSA (FLAIR)} 
\\
\cmidrule(lr){2-5} \cmidrule(lr){6-9} \cmidrule(lr){10-13} \cmidrule(lr){14-17} 

& 10\% & 30\% & 60\% & 100\% 
& 10\% & 30\% & 60\% & 100\% 
& 10\% & 30\% & 60\% & 100\% 
& 10\% & 30\% & 60\% & 100\% 
\\
\cmidrule(lr){2-17}

\multicolumn{17}{c}{\textbf{Dice Similarity Coefficient (DSC) (\%) $\bm{\uparrow}$}} \\
\midrule

\rowcolor{headergray}Finetune & 50.36 & 65.50 & 73.35 & \textbf{83.49} & 65.91 & 71.99 & 79.59 & 88.95 & 52.77 & 63.32 & 75.16 & 86.91 & 75.94 & 80.93 & 86.52 & 94.02\\
\midrule

L2P & 57.54 & 65.76 & 71.55 & 79.97 & 73.16 & 77.10 & 81.09 & 85.51 & 58.80 & 65.22 & 76.60 & 82.52 & 80.58 & 86.32 & 90.25 & 93.98\\

Moe-adapters & 57.77 & 66.15 & 72.64 & 80.26 & 74.35 & 77.31 & 82.52 & 85.77 & 61.96 & 71.64 & 80.16 & 83.53 & 80.33 & 86.96 & 90.46 & 93.09\\

DualPrompt & 58.24 & 64.95 & 74.41 & 82.92 & 75.92 & 78.31 & 83.85 & 88.17 & 59.35 & 66.70 & 76.71 & 85.66 & 82.86 & 85.52 & 91.32 & 94.50\\

CODA-P & 59.45 & 65.70 & 75.67 & 83.11 & 76.71 & 78.10 & 84.12 & 88.21 & 60.89 & 68.56 & 78.77 & 85.04 & 81.55 & 85.87 & 89.89 & 94.41\\

Low-Rank MoE & 61.77 & 67.12 & 75.36 & 82.26 & 76.47 & 79.88 & 84.61 & 88.77 & 61.74 & 71.64 & 81.60 & \textbf{86.99} & 80.33 & 86.96 & 92.80 & 94.70\\

SEMA & 61.16 & 66.70 & 76.02 & 82.40 & 77.41 & 80.53 & 84.03 & 89.78 & 62.52 & 73.59 & 82.80 & 85.31 & 82.93 & 89.45 & 93.73 & 94.46\\

Moe-adapters++ & 63.58 & 69.39 & 77.98 & 83.01 & 77.62 & 80.06 & 86.04 & 88.39 & 64.52 & 73.95 & 83.74 & 86.43 & 83.66 & 90.37 & 93.15 & 94.79\\

\rowcolor{color3} \textbf{CoRE (ours)} & \textbf{67.58} & \textbf{74.39} & \textbf{80.45} & 83.19 & \textbf{79.41} & \textbf{84.53} & \textbf{87.03} & \textbf{89.78} & \textbf{68.67} & \textbf{77.59} & \textbf{85.80} & 86.19 & \textbf{86.93} & \textbf{91.55} & \textbf{94.73} & \textbf{95.21}\\
\bottomrule[1.3pt]
\end{tabular}%
}
\end{small}
\end{table}

Table \ref{tab:few_shot_results} presents the comprehensive evaluation on the BraTS2023-SSA dataset. This downstream evaluation rigorously tests the models across four modalities under varying training data availability, specifically 10\%, 30\%, 60\%, and 100\%. 

The results indicate that CoRE demonstrates effective data utilization. Under conditions with highly limited data, our proposed method achieves competitive segmentation accuracy that is comparable to or exceeds other approaches trained with significantly more data. For instance, utilizing only 10\% of the training data on Task 15, CoRE attains a DSC of 68.67\%, which surpasses CODA-P using 30\% of the data (68.56\%). Similarly, on Task 16, CoRE with just 30\% data achieves 91.55\% DSC, outperforming DualPrompt trained on 60\% data (91.32\%). Furthermore, CoRE at 60\% data on Task 16 (94.73\%) exceeds the performance of CODA-P using the full 100\% dataset (94.41\%).

This transferability is attributed to the conceptual-reasoning framework. Methods relying purely on visual perception often struggle to generalize when data is scarce because they easily overfit to the limited visual samples. Conversely, CoRE leverages established clinical priors and conceptual alignments. By routing expert modules based on stable brain lesion attributes rather than fluctuating visual statistics, the model can quickly adapt to novel domains with minimal parameter updates. This mechanism ensures stable performance and rapid adaptation even when downstream task data is severely constrained.

\section{Further Extensive Analysis}
\label{analysis}

\begin{table}[t]
\centering
\caption{Performance evaluation of CoRE and comparative methods under a different task order (ATLAS $\rightarrow$ MSSEG $\rightarrow$ WMH $\rightarrow$ ISLES $\rightarrow$ BraTS) to analyze task order sensitivity. Bold font represents the best result (excluding offline JointTrain and task-specific IndividualTrain).}
\label{tab:task_order_sensitivity}
\begin{small}
\resizebox{\textwidth}{!}{%
\begin{tabular}{l c cccc cc c cccc c c}
\toprule[1.3pt]

\multirow{5}{*}{\textbf{Method}} 
& ATLAS 
& \multicolumn{4}{c}{MSSEG} 
& \multicolumn{2}{c}{WMH} 
& ISLES 
& \multicolumn{4}{c}{BraTS} 
& \multirow{5}{*}{\textbf{Average}} 
& \multirow{5}{*}{\textbf{BWT}} \\
\cmidrule(lr){2-2} \cmidrule(lr){3-6} \cmidrule(lr){7-8} \cmidrule(lr){9-9} \cmidrule(lr){10-13}

& T1w & T1w & T2w & T1ce & FLAIR & T1w & FLAIR & DWI & T1w & T2w & T1ce & FLAIR & & \\
\cmidrule(lr){2-2} \cmidrule(lr){3-6} \cmidrule(lr){7-8} \cmidrule(lr){9-9} \cmidrule(lr){10-13}

& Task 1 & Task 2 & Task 3 & Task 4 & Task 5 & Task 6 & Task 7 & Task 8 & Task 9 & Task 10 & Task 11 & Task 12 & & \\
\cmidrule(lr){2-15}

\multicolumn{15}{c}{\textbf{Dice Similarity Coefficient (DSC) (\%) $\bm{\uparrow}$}} \\
\midrule
\rowcolor{headergray} IndividualTrain & 71.28 & 70.81 & 72.49 & 70.50 & 80.91 & 76.77 & 83.62 & 87.98 & 85.78 & 90.35 & 86.35 & 92.81 & 80.80 & 100.00 \\
JointTrain & 70.31 & 70.62 & 73.63 & 70.18 & 79.82 & 75.23 & 81.98 & 86.01 & 84.78 & 90.18 & 87.02 & 92.67 & 80.20 & 99.19\\
Finetune & 0.00 & 0.09 & 8.39 & 0.00 & 55.12 & 2.90 & 60.78 & 71.77 & 62.33 & 73.22 & 77.49 & 93.29 & 42.11 & 57.75\\
\midrule

L2P & 56.96 & 53.45 & 57.39 & 44.23 & 73.28 & 67.36 & 77.38 & 81.22 & 80.23 & 86.79 & 82.35 & 90.23 & 70.91 & 89.44\\

Moe-adapters & 57.91 & 56.19 & 58.47 & 45.38 & 77.23 & 69.33 & 78.49 & 82.10 & 82.33 & 87.24 & 83.69 & 91.42 & 72.48 & 91.05\\

DualPrompt & 57.49 & 60.17 & 61.08 & 46.39 & 76.55 & 70.39 & 82.08 & 82.31 & 80.59 & 86.23 & 83.22 & 90.35 & 73.07 & 91.79\\

CODA-P & 62.34 & 62.92 & 61.75 & 46.23 & 78.96 & 73.49 & 83.63 & 83.28 & 81.25 & 86.49 & 81.44 & 91.34 & 74.43 & 93.18\\

Low-Rank MoE & 62.88 & 63.46 & 65.38 & 57.58 & 78.83 & 73.29 & 83.56 & 83.74 & 81.29 & 86.93 & 81.39 & 91.28 & 75.80 & 94.68 \\

SEMA & 64.50 & 65.78 & 68.67 & 62.35 & 80.23 & 74.01 & \textbf{83.82} & 85.79 & 82.86 & 88.93 & 83.91 & 92.14 & 77.75 & 96.71\\

Moe-adapters++ & 64.38 & 67.26 & 65.00 & 62.72 & \textbf{80.45} & 73.89 & 83.59 & 85.57 & 81.52 & 89.24 & 84.26 & 92.68 & 77.55 & 96.46\\

\rowcolor{color3} \textbf{CoRE (ours)} & \textbf{68.01} & \textbf{69.95} & \textbf{70.67} & \textbf{65.98} & 80.29 & \textbf{75.47} & 83.63 & \textbf{87.88} & \textbf{85.83} & \textbf{90.54} & \textbf{86.94} & \textbf{92.95} & \textbf{79.85} & \textbf{98.86}\\
\bottomrule[1.3pt]
\end{tabular}%
}
\end{small}
\end{table}

\subsection{Analysis of Task Order Sensitivity}
\label{Task-Order}
Continual learning models can be sensitive to the sequence of incoming data. To evaluate this stability, we altered the original task sequence by placing relatively challenging datasets (ATLAS and MSSEG) at the beginning of the data stream, forming a new training order: $\text{ATLAS} \rightarrow \text{MSSEG} \rightarrow \text{WMH} \rightarrow \text{ISLES} \rightarrow \text{BraTS}$. The comparative results for this new sequence are summarized in Table~\ref{tab:task_order_sensitivity}.

Approaches with a fixed expert pool, such as L2P and Moe-adapters, are more prone to severe forgetting in early tasks. This is because their limited capacity makes early representations easily overwritten by subsequent tasks. Conversely, other approaches remain largely unaffected by the sequence variation. For example, task-specific methods allocate strictly independent parameters for each new task, thereby preventing later learning from interfering with earlier experts. Similarly, image-perception methods and CoRE achieve this isolation by conditionally expanding their architecture and locking previously trained weights. When visual distribution shifts or joint conceptual-visual novelty are detected respectively, new capacity is allocated for the incoming data. Meanwhile, the parameters of existing modules remain strictly frozen, guaranteeing they can be safely reused for past tasks without being updated or overwritten. Consequently, our proposed framework maintains a consistent performance, achieving an average DSC of 79.85\% and a BWT of 98.86\% under the new sequence.

\subsection{Analysis of Expandable Layers}
\label{Expandable-Layers}
\begin{figure}[t!] 
    \centering     
    \includegraphics[width=\linewidth]{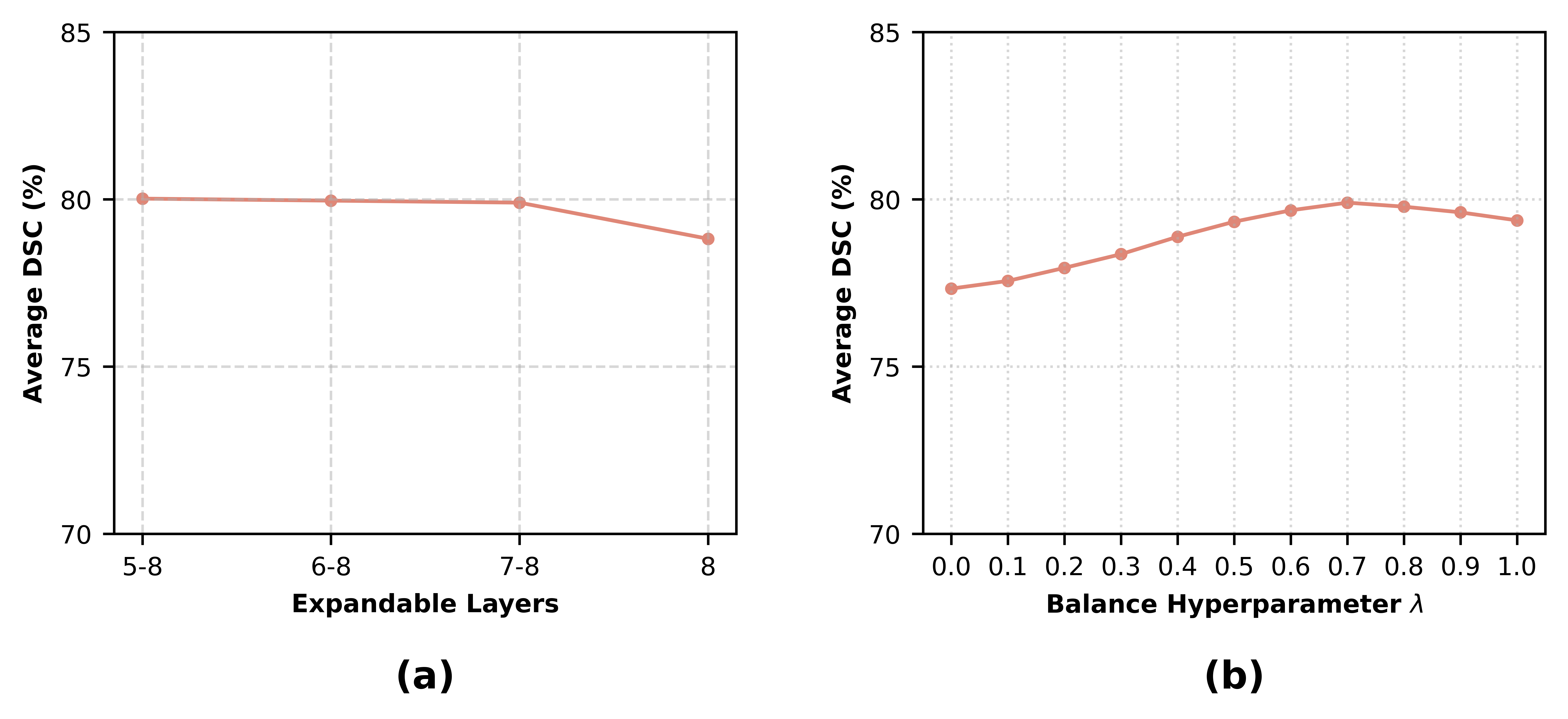} 
    \caption{Analysis of expandable layers and the balance hyperparameter $\lambda$. (a) Average DSC across the sequential tasks when dynamic expansion is applied to different combinations of transformer blocks. (b) Average DSC under varying values of the balance hyperparameter $\lambda$.}
    \label{fig:figure1}
\end{figure}

We investigate the impact of applying dynamic expansion to different sets of transformer blocks to determine the optimal architectural growth strategy. As illustrated in Fig.~\ref{fig:figure1} (a), allowing expansion in shallower transformer layers (such as blocks 5 and 6) does not meaningfully improve the overall segmentation performance. Shallow layers primarily capture low-level visual textures, which share similarities despite distribution shifts across different magnetic resonance imaging modalities. Consequently, introducing too many adapters in these early stages leads to severe parameter redundancy, making the minor performance fluctuations not worth the increased computational cost. Therefore, restricting the dynamic expansion to the final two transformer blocks (layers 7 and 8) effectively balances model capacity and segmentation accuracy. Restricting expansion to only the final block results in a noticeable performance drop, indicating that expanding only at the last level is insufficient to handle task heterogeneity.

\subsection{Analysis of Balance Hyperparameter}
\label{Balance-Hyperparameter}
We analyze the sensitivity of the balance hyperparameter $\lambda$ to understand its influence on expert routing. This parameter controls the fusion weight between the two routing signals. As shown in Fig.~\ref{fig:figure1} (b), a value of $\lambda=0$ indicates that the model relies entirely on the image side routing, resulting in the lowest average DSC. Conversely, a value of $\lambda=1$ represents a complete reliance on the concept side routing. While this yields better results than relying solely on images, it ignores the continuous distribution characteristics that discrete textual descriptions cannot fully capture. The best performance is achieved at $\lambda=0.7$, demonstrating that an appropriate combination of structured brain lesion concepts and visual feature distributions provides the most reliable guidance for expert selection.

\subsection{Analysis of Different LLM Generators}
\label{LLM-Generators}

\begin{table}[t]
\centering
\caption{Ablation study of different LLMs used for BLC-Lib generation.} 
\label{tab:ablation_results}
\begin{small}
\resizebox{\textwidth}{!}{%
\begin{tabular}{l cccc c cccc c cc c c}
\toprule[1.3pt]

\multirow{5}{*}{\textbf{Method}} 
& \multicolumn{4}{c}{BraTS} 
& ATLAS 
& \multicolumn{4}{c}{MSSEG} 
& ISLES & \multicolumn{2}{c}{WMH} 
& \multirow{5}{*}{\textbf{Average}} 
& \multirow{5}{*}{\textbf{BWT}} \\
\cmidrule(lr){2-5} \cmidrule(lr){6-6} \cmidrule(lr){7-10} \cmidrule(lr){11-11} \cmidrule(lr){12-13}

& T1w & T2w & T1ce & FLAIR & T1w & T1w & T2w & T1ce & FLAIR & DWI & T1w & FLAIR & & \\
\cmidrule(lr){2-5} \cmidrule(lr){6-6} \cmidrule(lr){7-10} \cmidrule(lr){11-11} \cmidrule(lr){12-13}

& Task 1 & Task 2 & Task 3 & Task 4 & Task 5 & Task 6 & Task 7 & Task 8 & Task 9 & Task 10 & Task 11 & Task 12 & & \\
\cmidrule(lr){2-13}

\multicolumn{15}{c}{\textbf{Dice Similarity Coefficient (DSC) (\%) $\bm{\uparrow}$}} \\
\midrule

Gemini 3 Pro & 85.52 & 89.98 & \textbf{86.97} & 92.34 & 68.04 & 70.52 & 70.33 & 66.22 & 79.79 & 87.48 & 76.45 & 83.57 & 79.77 & 98.82 \\
DeepSeek-V3.2 & 85.61 & 90.05 & 86.72 & 92.03 & 68.10 & 70.75 & \textbf{70.49} & 66.25 & 79.52 & 87.21 & 76.48 & 83.60 & 79.73 & 98.80 \\
Qwen3-235B-A22B & 85.59 & 90.02 & 86.84 & 92.30 & \textbf{68.19} & 70.78 & 70.38 & 66.31 & 79.81 & 87.49 & 76.42 & \textbf{83.72} & 79.82 & 98.87 \\
\rowcolor{color3}\textbf{GPT-5} & \textbf{85.64} & \textbf{90.11} & 86.89 & \textbf{92.87} & 68.15 & \textbf{70.83} & 70.41 & \textbf{66.37} & \textbf{79.86} & \textbf{87.54} & \textbf{76.51} & 83.66 & \textbf{79.90} & \textbf{98.96} \\

\bottomrule[1.3pt]

\end{tabular}%
}
\end{small}
\end{table}

In the main text, we utilize GPT-5 to construct the BLC-Lib. To investigate the impact of the chosen generator on the overall framework, we conduct a comprehensive analysis using different LLMs. We substitute GPT-5 with one advanced proprietary model (Gemini 3 Pro) and two state-of-the-art open-weight models (Qwen3-235B-A22B and DeepSeek-V3.2).

The quantitative results are presented in Table \ref{tab:ablation_results}. While GPT-5 is selected as the default generator in the main text due to its marginally superior average DSC and BWT, all evaluated LLMs demonstrate highly stable segmentation accuracy and knowledge retention. Specifically, the average DSC remains consistently between 79.73\% and 79.90\% across all models, and the BWT stays above 98.80\%.
These consistent results indicate that the CGC module is highly robust to textual variations in the generated concepts, suggesting that the effectiveness of the CoRE framework stems from its intrinsic concept-reasoning mechanism rather than a dependence on a specific proprietary model. Furthermore, achieving highly competitive performance with open-weight models not only validates the generalizability of the framework but also highlights its practical viability for secure local deployment in clinical environments governed by strict data privacy regulations.

\end{document}